\PassOptionsToPackage{dvipsnames,table,xcdraw}{xcolor}
\documentclass[acmtog,nonacm]{acmart}
\usepackage{multicol}
\usepackage{multirow}

\usepackage{pgfplotstable}
\usepackage{pgffor}
\usepackage{ifthen}
\usepackage{booktabs}
\usepackage{makecell}
\usepackage{listings}
\usepackage{graphicx}
\usepackage{calc}
\usepackage{wrapfig}
\usepackage{afterpage}
\usepackage{tablefootnote}

\usepackage{pgfplots}
\pgfplotsset{compat=1.18}

\newcommand{\methodname}{SEIG}

\newcommand{\vigafull}{VIGA\textsuperscript{full}}
\newcommand{\vigaonly}{VIGA\textsuperscript{VLM-only}}

\makeatletter
\newcommand\myparagraph{\@startsection{paragraph}{4}{\parindent}%
  {1sp}%
  {-1.5\p@}%
  {\ACM@NRadjust{\@parfont\@adddotafter}}}
\makeatother

\begin{document}
\title{Thinking in Blender: Staged Executable Inverse Graphics with Vision-Language Models}

\author{Guangzhao He}\authornote{Denotes equal contribution.}
\affiliation{%
  \institution{Cornell University}
  \country{USA}
}
\email{gh466@cornell.edu}

\author{Rundong Luo}\authornotemark[1]
\affiliation{%
  \institution{Cornell University}
  \country{USA}
}
\email{rl897@cornell.edu}

\author{Wei-Chiu Ma}\authornote{Denotes equal advising.}
\affiliation{%
  \institution{Cornell University}
  \country{USA}
}
\email{wm347@cornell.edu}

\author{Hadar Averbuch-Elor}\authornotemark[2]
\affiliation{%
  \institution{Cornell University}
  \country{USA}
}
\email{hadarelor@cornell.edu}

\begin{abstract}
Inverse graphics is a longstanding and highly underconstrained problem that seeks to reconstruct images as editable 3D scenes which can be rendered, relit, and manipulated. In this work, we investigate whether pretrained vision-language models (VLMs) can perform executable inverse graphics directly from a single image by reconstructing a scene as an editable Blender program, without relying on specialized 2D or 3D foundation models, differentiable rendering, or multi-view supervision. We introduce \textbf{S}taged \textbf{E}xecutable \textbf{I}nverse \textbf{G}raphics (\methodname{}), an agentic framework that reconstructs a 3D scene from a single image by progressively refining scene factors including geometry, materials, composition, and lighting directly in executable Blender code space. We evaluate our framework across diverse scenes using a range of reconstruction metrics spanning pixel-level, perceptual, and semantic fidelity. Our experiments show that staged reconstruction substantially improves reconstruction fidelity, highlighting the importance of task decomposition for executable inverse graphics with general-purpose VLMs. Finally, we showcase various downstream applications enabled by the reconstructed editable Blender scenes.

\end{abstract}
\begin{teaserfigure}
  \centering
  \includegraphics[width=\textwidth]{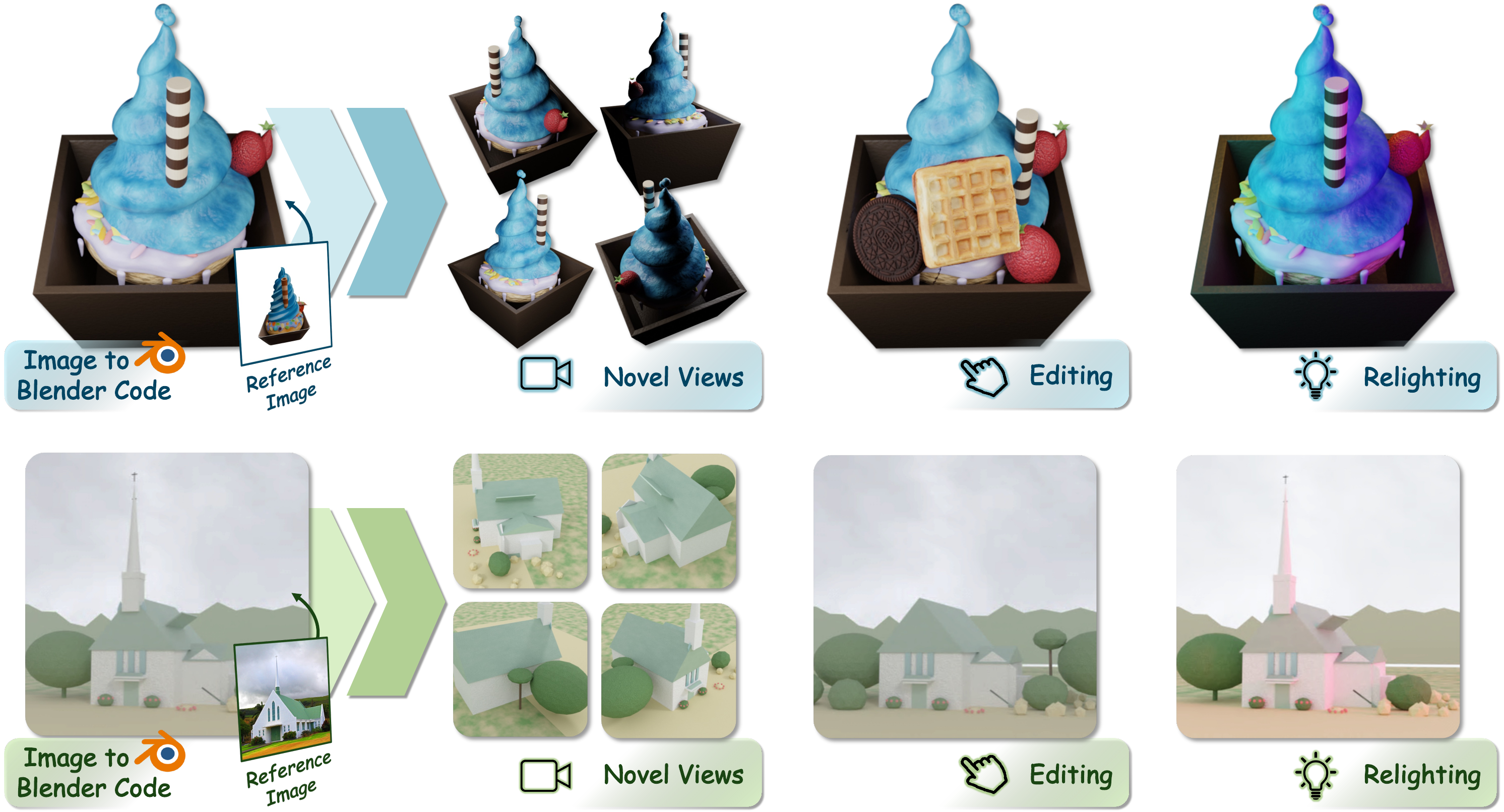}
  \vspace{-15pt}
  \caption{ From a single reference image (leftmost inset), \methodname{} reconstructs an editable Blender scene through a staged generator--verifier loop driven entirely by a pretrained VLM.
  As the output is a structured Blender program, our approach directly supports \emph{novel-view synthesis}, \emph{editing}, and \emph{relighting} (right). }%
  \Description{Overview diagram of an inverse graphics pipeline where image inputs are processed by a vision-language-model agent that generates Blender code for scene reconstruction and downstream applications.}
  \vspace{6pt}
\end{teaserfigure}

\maketitle

\section{Introduction}

Creating 3D scenes is a complex and labor-intensive process that requires extensive expertise in specialized graphics software such as Blender. Professional artists typically construct scenes through a staged but highly iterative workflow: modeling or assembling geometry, assigning materials and textures, arranging scene composition, configuring lighting, and carefully adjusting camera parameters. Throughout this process, artists continuously inspect intermediate results and iteratively refine individual scene factors until the final scene matches the desired visual target. Replicating this capability automatically from images has long been a central goal of inverse graphics~\cite{roberts1963machine,barrow1978recovering,marschner1998inverse}.

Recent advances in vision-language models (VLMs) have demonstrated remarkable capabilities in visual reasoning, instruction following, and code generation~\cite{liu2023daxiespromptsunleashing3dspatial,hu2024scenecraft,yin2026viga}, suggesting that these models may encode rich latent knowledge about 3D scenes and their underlying structure. In this work, we investigate whether pretrained VLMs can perform executable inverse graphics directly from a single image. Specifically, we ask whether VLMs can reconstruct a scene as an editable Blender program by recovering its underlying factors---including geometry, materials, composition and lighting---without relying on specialized 2D or 3D foundation models, differentiable rendering pipelines, or large-scale multi-view supervision.

Our key observation is that, while pretrained VLMs struggle to reconstruct all scene factors simultaneously, their executable inverse graphics capabilities can be unlocked by decomposing reconstruction into sequential, semantically meaningful stages that mirror the iterative workflow used by professional 3D artists. Building on this observation, we introduce \methodname{}, a \textbf{S}taged \textbf{E}xecutable \textbf{I}nverse \textbf{G}raphics framework built on top of a pretrained vision-language model. Starting from a single input image, \methodname{} first initializes a coarse scene scaffold composed of simple geometric primitives and approximate object layouts, and then progressively refines this representation through sequential stages that recover geometry, materials, object composition and lighting directly in executable Blender code space. Each stage is paired with a verifier module that renders and evaluates the current scene state before guiding subsequent refinements. By decomposing inverse graphics into sequential executable refinement stages, our framework reduces the complexity of the overall reconstruction problem while maintaining a fully editable and physically grounded scene representation throughout the reconstruction process.

We demonstrate our approach across both synthetic and in-the-wild scenes, comparing against monolithic executable inverse graphics baselines with and without specialist 2D and 3D foundation models. Our experiments show that staged reconstruction substantially improves reconstruction fidelity, suggesting that task decomposition may play a more critical role than the richness of the external toolkit used by the pipeline and that pretrained vision-language models encode surprisingly rich latent priors about 3D structure, appearance, and scene composition. Finally, we demonstrate the applicability of our approach to downstream graphics tasks including relighting, scene editing and physics simulation, enabled directly through the reconstructed editable Blender scene representation.

\section{Related Work}
\label{sec:related_works}

 \noindent
\textbf{Inverse Graphics}. Recovering structured 3D scenes from 2D images has long been a central goal in computer vision and graphics, dating back to early formulations of inverse graphics such as Roberts' ``Blocks World''~\cite{roberts1963machine}. Early works in inverse graphics primarily focused on the inverse rendering problem, seeking to recover scene geometry, illumination, and reflectance from images through analysis-by-synthesis formulations~\cite{marschner1998inverse,ramamoorthi2001signal}, intrinsic decomposition~\cite{barrow1978recovering} and shape-from-shading~\cite{horn1970shape}. More recent approaches extend these ideas to recovering geometry and reflectance from sparse views or single images using neural networks~\cite{dong2014appearance,nam2018practical,li2018learning,bi2020deep}, while several works further explore structured scene decomposition through differentiable rendering and primitive-based representations~\cite{sharma2018csgnet,monnier2023differentiable}.

In recent years, inverse graphics research has increasingly shifted toward neural scene representations such as NeRF~\cite{mildenhall2020nerf} and 3D Gaussian Splatting~\cite{kerbl2023gaussiansplatting}. While these neural representations are highly effective for scene reconstruction, they typically encode geometry, materials, and lighting in latent neural representations that are not directly editable as structured graphics programs. Several works seek partial disentanglement within this paradigm by factoring shape and reflectance~\cite{zhang2021nerfactor,zhang2021physg}, separating geometry, BRDFs, and illumination~\cite{jin2024tensoir,munkberg2022nvdiffrec,liang2024gsir},
or introducing object-level scene decomposition~\cite{yang2021learning,wu2024dynamic,benaim2024volumetric,luo2025unsupervised}. Despite this progress, these methods still do not directly recover executable scene programs.

\smallskip \noindent
\textbf{Vision-Language Models for 3D Reasoning.} Vision-language models (VLMs) demonstrate strong capabilities in visual understanding, instruction following, and code generation~\cite{liu2023llava,openai2023gpt4v,team2023gemini}. Beyond semantic reasoning, several works show that these models also encode non-trivial spatial and geometric understanding from 2D observations. Prior studies demonstrate that, with appropriate prompting, VLMs can perform coarse 3D grounding, spatial reasoning, and geometric estimation from images~\cite{liu2023daxiespromptsunleashing3dspatial}, while benchmarks such as SpatialVLM~\cite{chen2024spatialvlm} systematically evaluate these capabilities across diverse spatial tasks. At the same time, recent evaluations reveal that current VLMs remain significantly stronger at semantic reasoning than at precise geometric prediction, particularly in tasks requiring accurate spatial localization or metric 3D understanding~\cite{kulits2024rethinkinginversegraphicslarge}. These observations motivate our investigation into whether pretrained VLMs can nevertheless support executable inverse graphics from a single input image through staged reconstruction and iterative visual verification.

\smallskip \noindent
\textbf{Executable Scene Generation with Vision-Language Models.} Recent works explore using vision-language models to generate and manipulate 3D content through executable scene representations. SceneCraft~\cite{hu2024scenecraft} and LL3M~\cite{lu2025ll3m} investigate generating executable 3D scenes as Blender programs from text instructions. MeshCoder~\cite{dai2025meshcoder} instead targets code-based mesh generation from point clouds, while BrickGPT~\cite{pun2025brickgpt} explores compositional 3D generation through structured primitive-based representations. Articulate-Anything~\cite{le2024articulate} and VDAWorld~\cite{omahony2025vdaworld} further extend these ideas to generating articulated assets and simulation-ready environments from multimodal inputs. BlenderGym~\cite{gu2025blendergym} and IR3D-Bench~\cite{liu2025ir3dbench} focus on evaluating VLM-driven 3D editing and reconstruction systems, highlighting geometric precision and spatial consistency as major limitations of current models.

Closest to our work, VIGA~\cite{yin2026viga} formulates vision-as-inverse-graphics as an iterative write--render--compare--revise loop, enabling executable 3D scene reconstruction from multimodal inputs, including the single image setting addressed in our work. As illustrated in our experiments, however, treating reconstruction as a single monolithic optimization problem remains highly challenging for current VLMs, which struggle to jointly reason about geometry, materials, composition and lighting,  within a single entangled generation process. In contrast, our framework explicitly decomposes reconstruction into sequential generator--verifier stages that independently recover individual scene factors. Furthermore, unlike prior work, our approach operates entirely using a single pretrained VLM without relying on specialized 2D or 3D foundation models, such as SAM~\cite{kirillov2023sam} and SAM-3D~\cite{meta2025sam3d}. This design allows us to more directly investigate the extent to which pretrained VLMs themselves encode the geometric, spatial, and compositional priors necessary for executable inverse graphics.

\begin{figure*}
  \centering
  \includegraphics[width=\textwidth]{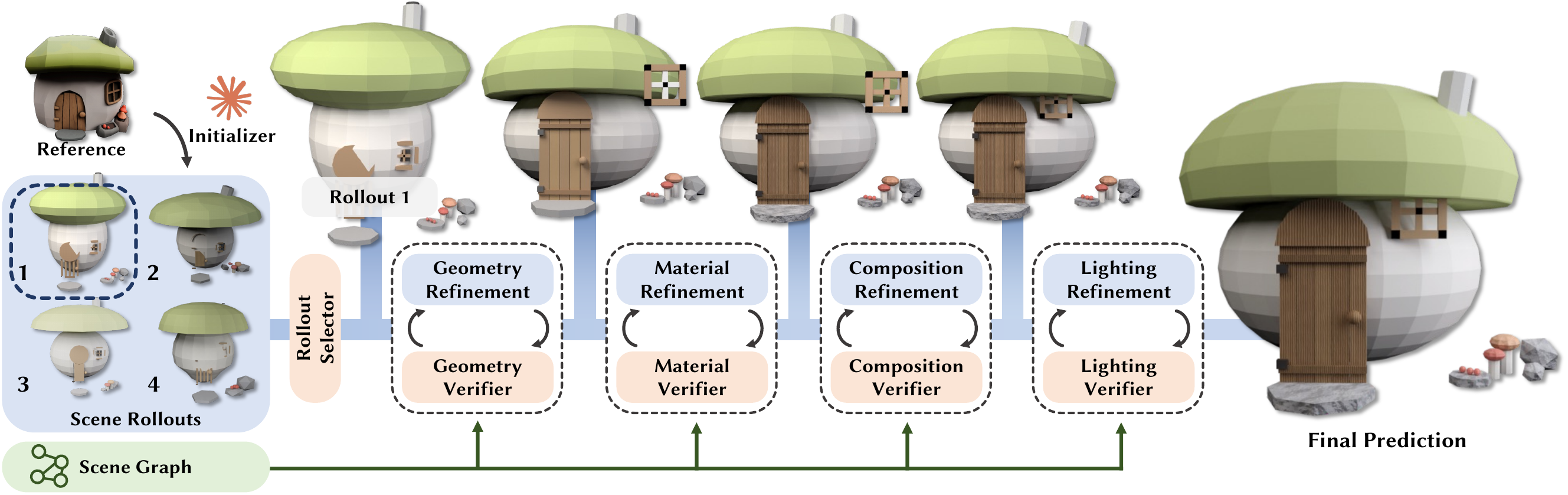}
  \vspace{-15pt}
  \caption{\textbf{Method overview.}
  Our agentic reconstruction pipeline decomposes inverse graphics into four sequential stages of Blender code generation, each consisting of a \textit{generator} step followed by a \textit{verifier} step.
  An initialization stage samples four coarse scene initializations and labels a scene graph for all objects and parts, which is then passed to all subsequent stages for object- and part-centric refinement.
  The scene graph records each object with attributes such as geometry, material, and spatial relations.
  The final scene is produced by executing the refined Blender code, reconstructing the reference image.
  }
  \Description{Flowchart of the proposed staged inverse graphics pipeline with generator and verifier steps for geometry, material, composition, and lighting, followed by final rendering.}
  \vspace{-5pt}
  \label{fig:method}
\end{figure*}

\section{Method}

In this section, we introduce \methodname{}: a framework that takes a reference image as input, reconstructs the underlying scene by decomposing the problem into stages, and produces executable Blender code whose render matches the input image.
We use Blender as the reconstruction engine since it provides a unified Python interface for scene editing and image rendering, allowing the agent to modify the scene and observe it via simple API calls.
Rather than requiring hours of manual construction and refinement by an experienced artist, \methodname{} instead delegates this workflow to a vision-language model (VLM) that interprets the input image, reasons about scene structure, and interacts with Blender through tool calls.

However, directly prompting a VLM to generate or refine a complete Blender scene places a heavy burden on the model, requiring it to jointly infer geometry, materials, composition, and lighting parameters while producing executable code. 
Prior VLM-based inverse graphics agents~\cite{yin2026viga,gu2025blendergym} demonstrate the promise of this direction, but still leave many coupled scene factors to be solved at once.
In this work, we address this by decomposing inverse graphics into independently scoped, verifiable subproblems.
Central to our approach is a multi-stage, additive pipeline in which the scene is constructed and refined autoregressively.
Each stage is disentangled to depend only on earlier ones (e.g., geometry before material refinement, material refinement before lighting), so the framework can commit to each subproblem independently before proceeding (Sec.~\ref{sec:stage}).
Within each stage, a generator-verifier loop drives iterative refinement: the generator writes Blender code across multiple tool-use rounds, then passes the rendered result to the verifier, which decides whether to request a revision or advance to the next stage (Sec.~\ref{sec:verifier}).
An overview is shown in Fig.~\ref{fig:method}.

\subsection{Staged Scene Construction}
\label{sec:stage}
Inverse graphics requires solving a set of tightly coupled subtasks, including reconstructing object geometry and material appearance, predicting the scene layout, and recovering environment lighting.
With each of these subtasks being challenging even for human artists or specialist models~\cite{verbin2022refnerf,jin2024tensoir,gao2024cat3d,wang2025vggt,zhang2024scenelanguage}, asking a VLM agent to directly generate code for the entire scene is therefore a highly underconstrained problem.
While prior works~\cite{yin2026viga,gu2025blendergym} attempt to generate and refine the entire scene at once, we argue that optimizing all such factors jointly creates a large search space in which errors in one factor can obscure corrections to another.
To this end, we propose a greedy, staged approach, in which the problem is decomposed into verifiable substages, enabling more grounded reasoning, and allowing the VLM to additively construct the scene through disentangled steps.

More specifically, we follow the conventional workflow used by human artists when creating renders in Blender: initializing the scene, modeling individual objects, drawing texture maps and assigning materials, composing the objects into a scene, and finally adding environment lighting.
We formulate each stage as an agentic function that depends only on the outputs of previous stages while maintaining its own stage-specific context for reasoning.

\smallskip\noindent\textbf{Scene decomposition.}
Given a reference image, we first prompt the VLM to decompose the scene into a hierarchical scene graph that covers all visible objects.
The graph contains a scene root for the global environment and object nodes for physical objects or object parts.
Each node stores a visual description, approximate geometry, material appearance, spatial relations to its parent and nearby nodes, and a Blender reconstruction strategy.
The VLM then recursively refines the graph until each leaf node corresponds to an atomic component that can be approximated with Blender primitives (e.g., spheres, cubes or cones).
This representation encourages full scene coverage and provides later stages with referable object names for localized refinement.

\smallskip\noindent\textbf{Scene initialization.}
Given the scene graph, the scene initialization stage instantiates a coarse Blender scaffold containing all components in the graph.
The VLM first produces an execution plan from the scene-graph attributes, then generates Blender code that creates each leaf node with an initial geometry and material approximation according to the plan.
This scaffold, built only from textual descriptions, is not intended to match the reference image precisely, but to ensure that every decomposed component exists in the scene and is assigned a stable Blender object name across stages, providing a consistent reference throughout.
During this process, it also initializes the lighting and camera coarsely to ensure that all scene components are clearly
visible without cropping or over-exposure.

Since scene initialization determines the decomposition and coarse scaffold used by all later stages, failure to cover all objects or bad association can be difficult to recover from local refinement alone.
We therefore use a rollout sampling strategy during initialization: we sample multiple independent scene graphs and coarse Blender scaffolds, then apply a rollout selector to select the candidate with the most complete object coverage and most plausible structure.
The selected rollout is then passed to the remaining stages, where iterative refinement optimizes different factors in sequence.

\smallskip\noindent\textbf{Geometry stage.}
After initialization produces a coarse scaffold, in which each scene graph leaf node is instantiated as a named Blender object, the geometry stage refines the shape of each individual object.
More specifically, a VLM is asked to refine each object's shape and physical structure through three classes of edits: (1) local shape edits, such as adjusting meshes and curves; (2) geometric transforms, such as scaling, rotating, and aligning existing objects parts; and (3) structural edits, such as adding missing parts or organizing object's internal hierarchies.
To support object-centric refinement, we provide the VLM with interaction tools for inspecting and editing the scene. 
The agent can call them to render the scene from alternative viewpoints, isolate individual objects for focused inspection, and revert unsuccessful edits when visual feedback indicates a regression.
The result from this stage is a geometrically refined scene with proper object identities, providing the structural foundation needed for subsequent material refinement and scene-level composition.

\smallskip\noindent\textbf{Material stage.}
After object geometry is refined, the material stage completes the object-wise reconstruction by matching each object's material and surface appearance with the reference image.
While the initialization may provide coarse, often single-color textures, the material stage replaces these placeholders with more detailed Blender PBR materials.
For each object in the scene graph, the agent initializes material slots and UV maps when needed, then creates procedural or image textures through Blender shader nodes.
The agent edits surface properties such as base color, roughness, specular, metallic, alpha, and bump or normal maps, capturing material identity and local texture detail.
To prevent material refinement from altering earlier stage results, the model executes Blender code through a material-only tool that permits only material-related edits.

\smallskip\noindent\textbf{Composition stage.}
After object-level geometry and material refinement, the composition stage arranges the finalized objects to match the reference image at the scene layout level.
Unlike the previous object-centric stages, this stage compares the target-view render against the reference and adjusts object transforms to match their relative scale, position, rotation, contact, and overall spatial organization.
During this process, the VLM agent may adjust the target-view camera when necessary to compare the scene with the reference, and may freely use temporary arbitrary-view renders to judge layout from other viewpoints.
However, the agent is not allowed to edit object geometry or materials.

\smallskip\noindent\textbf{Lighting stage.}
After geometry, material, and composition refinement, the lighting stage serves as the final scene-level adjustment, matching the rendered appearance of the accumulated Blender scene to the reference image by optimizing lighting parameters while keeping object shape, appearance, layout, and camera fixed.
The lighting stage agent compares the current render with the reference image to infer lighting cues such as light direction and height, shadow direction and softness, color temperature, exposure, and contrast. 
The agent can adjust both the physical lighting controls, including light type, position, direction, energy, color, size, softness, and ambient lighting, as well as render settings such as exposure and color management.
Since lighting parameters are sensitive to small changes, the agent is instructed to make conservative edits and revert changes that make the render too dark or overexposed.

\subsection{Intra-Stage Generator-Verifier Refinement}
\label{sec:verifier}

While our staged pipeline decomposes inverse graphics into simpler subproblems, each stage still requires iterative refinement since a single code-generation pass is insufficient for matching the reference image.
Therefore, we adopt a  multi-round generator-verifier loop per stage, where the generator calls tools to inspect the current Blender scene, writes stage-specific code, executes the edit, and renders the updated result.
After each generator attempt, a verifier compares the rendered image against the reference and inspects the  scene through its own set of tools to identify remaining stage-specific mismatches.
Crucially, each verifier is scoped to its corresponding stage: it judges only the active factor (\emph{e.g.}, object presence for
initialization), while ignoring errors assigned to other stages.

Free-form verifier critiques can be noisy across attempts, giving the generator inconsistent targets and preventing convergence. 
We therefore require the verifier to return an explicit approval checklist: a concrete, actionable todo list of visual discrepancies that is injected into the generator context for the next attempt, and once the generator attempt satisfies these conditions, the verifier must approve it to move on to the next stage.
To prevent the refinement effectiveness from degrading over time due to accumulated context, we impose a stage-specific maximum round budget.
If each generator-verifier loop reaches its budget without satisfying the checklist, the verifier then must select the best attempt to move forward to the next stage. 
We set the round budget according to the complexity of each stage.
In practice, we allow five rounds for geometry refinement, three rounds each for material and composition refinement, and two rounds for lighting refinement.

\section{Experiments}

In this section, we describe our main experimental results. Specifically,
we seek to answer the following questions:
\begin{itemize}
    \item Can a staged executable inverse graphics framework, built entirely on top of an off-the-shelf VLM, produce structured Blender reconstructions that faithfully recover geometry, appearance, layout, and lighting from a single image?
    \item  How much of the reconstruction quality stems from the staged reconstruction framework itself, particularly when compared against existing executable inverse graphics systems both with and without specialized external tools?
    \item Are the resulting scenes genuinely useful as graphics assets, supporting standard downstream operations such as novel-view synthesis, relighting, and object editing without any further training?
\end{itemize}

In what follows, we first describe our experimental setup (Section \ref{sec:setup}). We then present quantitative and qualitative comparisons and results (Sections~\ref{sec:quant} and~\ref{sec:qualitative}, respectively). Finally, we showcase the downstream applications enabled by the reconstructed scene representation (Section~\ref{sec:applications}) and discuss limitations (Section~\ref{sec:limitations}). In the supplementary material, we provide full prompts used by our systems along with chat history for a sample input image.

\subsection{Setup}
\label{sec:setup}

\myparagraph{Implementation Details.}
We use Claude Opus~4.7~\cite{anthropic2025claude}, accessed through the Anthropic API, as the base VLM for both the generator and the verifier at every stage of our pipeline.
The same model is used throughout all experiments without any fine-tuning, prompt-tuning, or task-specific supervision, so any observed difference in reconstruction quality between methods can be attributed to harness design rather than the underlying model.

\myparagraph{Baselines and Metrics.}
We compare our framework with VIGA~\cite{yin2026viga}, the closest monolithic agentic baseline for executable inverse graphics, in two configurations.
\vigafull{} is the original VIGA pipeline, which uses SAM~\cite{kirillov2023sam} and SAM-3D~\cite{meta2025sam3d} to pre-segment and pre-reconstruct individual objects before invoking its agentic write--render--compare--revise loop.
\vigaonly{} is an ablation of VIGA that disables these specialist 2D and 3D foundation models, leaving only the VLM-driven agentic loop.
Reporting both isolates how much of VIGA's performance comes from its specialist toolchain versus its VLM agent, and makes the comparison with our pipeline---which also relies on the VLM alone---an apples-to-apples test of harness design.
For a fair comparison, all methods share the same backbone VLM.

For quantitative evaluation, we curate a set of images from the NeRF synthetic dataset~\cite{mildenhall2020nerf} and VoxHammer~\cite{li2025voxhammer}. Five images are rendered from 7 of the 8 NeRF scenes; the \emph{materials} scene containing specular reflections from metallic spheres is excluded from our evaluation. Likewise, we gather 13 object-centric scenes from~\cite{li2025voxhammer}. We report six metrics between each reconstructed rendering and its reference image: PSNR and SSIM at the pixel level; LPIPS and DreamSim~\cite{fu2023dreamsim} as learned perceptual scores; and two semantic similarities, DINO~\cite{oquab2024dinov2} (cosine similarity between [CLS]-token features of a DINOv2 ViT-L/14 encoder) and CLIP~\cite{radford2021clip} (cosine similarity between image embeddings from a CLIP ViT-B/32 encoder).
When reference meshes are available, we avoid conflating reconstruction quality with the agent's camera estimate by registering each reconstructed mesh to the reference by running both Neural Deformation Pyramid (NDP)~\cite{li2022ndp} and ICP and keeping whichever yields the smaller Chamfer distance, then rendering the aligned scene from the reference camera and computing the six metrics on that rendering.  Additionally, we collect a set of images to stress-test the framework on in-the-wild scenarios.

\begin{table}
  \centering
  \small
  \setlength{\tabcolsep}{4pt}
  \renewcommand{\arraystretch}{1.2}
  \scalebox{0.91}{
  \begin{tabular}{l|ccc|ccc}
    \hline
    Method & PSNR $\uparrow$ & SSIM $\uparrow$ & LPIPS $\downarrow$ & DreamSim $\downarrow$ & DINO $\uparrow$ & CLIP $\uparrow$ \\
    \hline
    \vigaonly & 12.33 & \textbf{0.7122} & 0.3506 & 0.3693 & 0.6221 & 0.8451 \\
    \vigafull & 11.18 & 0.6647 & 0.3944 & 0.3624 & 0.5545 & 0.7986 \\
    \textbf{Ours} & \textbf{13.58} & 0.6881 & \textbf{0.3493} & \textbf{0.3021} & \textbf{0.7188} & \textbf{0.8830} \\
    \hline
  \end{tabular}
  }
  \vspace{1mm}
  \caption{\textbf{Quantitative comparison on NeRF synthetic scenes.} PSNR, SSIM, LPIPS, DreamSim, DINO, and CLIP between each method's reconstructed rendering and the reference image.
  }
  \label{tab:nerf-results}
\end{table}

\begin{table}
  \centering
  \small
  \vspace{-5mm}
  \setlength{\tabcolsep}{3pt}
  \renewcommand{\arraystretch}{1.2}
  \resizebox{\columnwidth}{!}{
  \begin{tabular}{l|ccc|ccc}
    \hline
    Method & PSNR $\uparrow$ & SSIM $\uparrow$ & LPIPS $\downarrow$ & DreamSim $\downarrow$ & DINO $\uparrow$ & CLIP $\uparrow$ \\
    \hline
    \vigaonly & 11.52 & \textbf{0.6776} & 0.3931 & 0.3847 & 0.5606 & 0.8366 \\
    \vigafull & 12.48 & 0.6743 & 0.4466 & 0.4441 & 0.4832 & 0.7883 \\
    \textbf{Ours} & \textbf{12.65} & 0.6737 & \textbf{0.3823} & \textbf{0.3433} & \textbf{0.6293} & \textbf{0.8446} \\
    \hline
  \end{tabular}
  }
  \vspace{1mm}
  \caption{\textbf{Quantitative comparison on Edit3D.} We report the same metrics as in Tab.~\ref{tab:nerf-results} over scenes gathered from~\cite{li2025voxhammer}.}
  \vspace{-6.5mm}
  \label{tab:edit3d-results}
\end{table}

\subsection{Quantitative Results}
\label{sec:quant}
Tab.~\ref{tab:nerf-results} and Tab.~\ref{tab:edit3d-results} report the quantitative comparison on the NeRF synthetic and Edit3D sets.
\methodname{} achieves the best score on five out of six metrics on both the NeRF synthetic and Edit3D scenes.
Despite using no specialist 2D or 3D foundation models, \methodname{} outperforms \vigafull{}, indicating that the gains come from harness design rather than tool access; that it also outperforms \vigaonly{} verifies the contribution of our per-stage decomposition.
This is consistent with the finding from BlenderGym~\cite{gu2025blendergym} and IR3D-Bench~\cite{liu2025ir3dbench} that visual precision, not tool orchestration, is the dominant bottleneck in current agentic 3D pipelines.

\begin{figure}
    \centering
    \vspace{1mm}
    \includegraphics[width=0.99\linewidth]{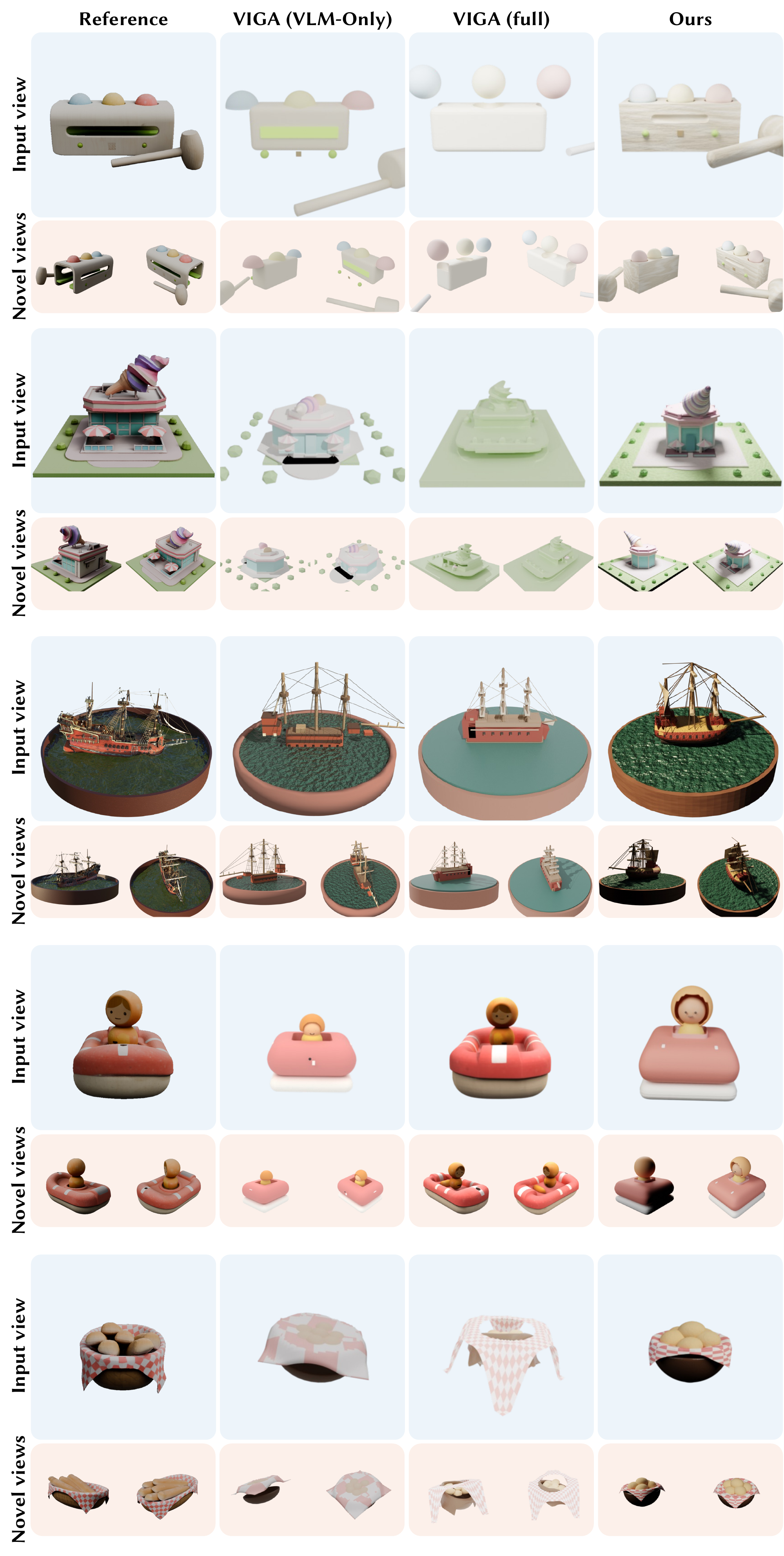}
    \vspace{-2mm}
    \caption{\textbf{Qualitative comparison across methods.}
    Each block shows a different reference image (leftmost column) together with the reconstructions produced by \vigaonly{}, \vigafull{}, and our pipeline.
    Within each method's column, the large rendering is taken from the reference viewpoint, and the two smaller show the same reconstructed scene rendered from alternate viewpoints, revealing the underlying 3D structure.
    Across these scenes, our pipeline reproduces the global geometry, materials, and object composition of each reference despite using no specialist 3D foundation models; \vigaonly{} recovers a plausible silhouette but loses smaller scene elements and surface detail, while \vigafull{} often produces fragmented, mis-colored meshes that fail to assemble into a coherent scene since the VLM agent may overwrite the texture of the 3D objects generated by SAM-3D.
    }
    \Description{Qualitative comparison across methods.}
    \label{fig:qualitative-side-by-side}
\end{figure}

\subsection{Qualitative Results}
\label{sec:qualitative}
\begin{figure*}[!p]
    \centering
    \begin{minipage}[c][\textheight][c]{\linewidth}
        \centering
        \includegraphics[width=0.98\linewidth]{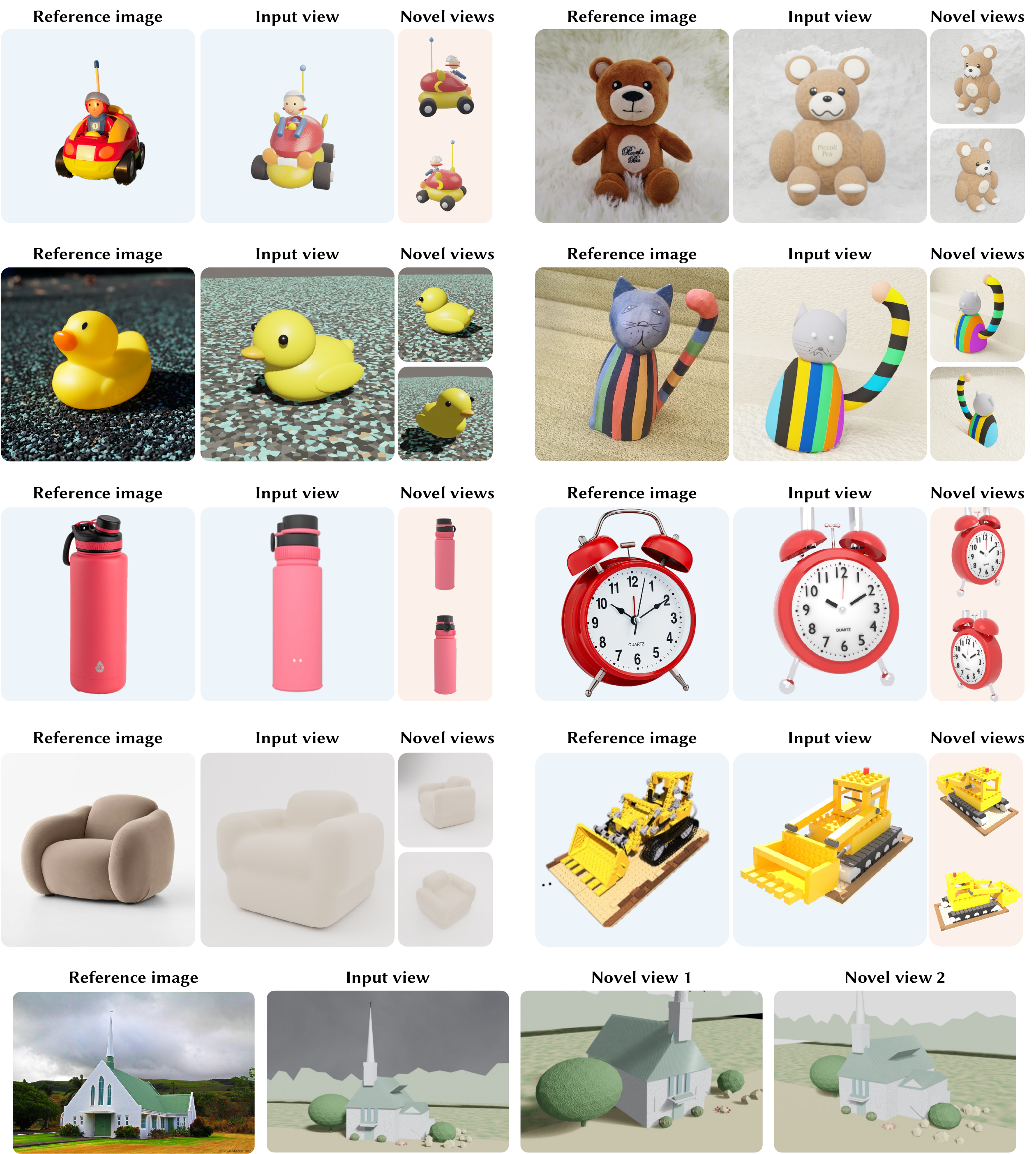}
        \caption{Gallery of Blender scenes created by \methodname{} from in-the-wild and synthetic reference images. The  synthetic scenes correspond to the examples shown in the first row (left) and the fourth row (right). Input and novel views are visualized along with the input reference image. }
        \Description{Gallery 1}
        \label{fig:gallery}
    \end{minipage}
\end{figure*}

Fig.~\ref{fig:gallery} shows representative reconstructions produced by our pipeline. Across these scenes, our framework produces structured Blender outputs that match the reference images in geometry, surface appearance, and composition, demonstrating that careful harness design alone, without any task-specific training, is sufficient to produce high-quality inverse graphics from an off-the-shelf VLM.
Fig.~\ref{fig:qualitative-side-by-side} compares these reconstructions with both VIGA variants; our method produces more accurate reconstructions than either configuration across geometry, material, and composition on most cases, while \vigaonly{} loses smaller scene elements and \vigafull{} fragments into disjoint, mis-colored meshes, as the underlying VLM agent often overwrites the texture of the generated 3D objects.

As can be observed in the fourth example of Fig.~\ref{fig:qualitative-side-by-side} (the humanoid character), even a strong single-image 3D generator such as SAM-3D may lead to characteristic single-view lifting artifacts. In this case, \vigafull{} exhibits the well-known \emph{Janus} failure mode, where frontal facial features are duplicated onto the back of the character's head due to ambiguity in the unseen object regions. In contrast, \vigaonly{} and our pipeline avoid this failure mode by reconstructing the figure compositionally from primitives rather than lifting a single-view mesh prior.
The bread-basket scene in the bottom of Fig.~\ref{fig:qualitative-side-by-side} illustrates the inherently underdetermined nature of single-view reconstruction: the reference's contents are mostly occluded, and although the ground truth is a basket of bread sticks, our pipeline instead produces rounded loaves---an interpretation equally consistent with the visible silhouette and rendered as a perceptually coherent scene.
Both VIGA variants, by contrast, fail to recover even a coherent basket structure on the same input, indicating that their failure mode here is not view ambiguity but a lack of compositional discipline in the monolithic agentic loop.

Beyond the final reconstructions, Fig.~\ref{fig:intermediate} traces the pipeline's intermediate outputs through each stage on two example scenes, demonstrating the importance of our staged approach. Starting from a coarse primitive-based scaffold, the four stages progressively refine the scene: geometry refinement sharpens individual object shapes, material assignment adds surface details, composition places the objects in their reference layout, and lighting configures the illumination. The final image is then rendered from a VLM-determined camera. Because each stage commits its output before the next begins, every intermediate scene itself is a coherent, editable Blender program---a property that distinguishes staged reconstruction from monolithic agentic pipelines and makes the pipeline's decisions inspectable and selectively reusable at any stage boundary.

\begin{figure}
    \centering
    \includegraphics[width=\columnwidth]{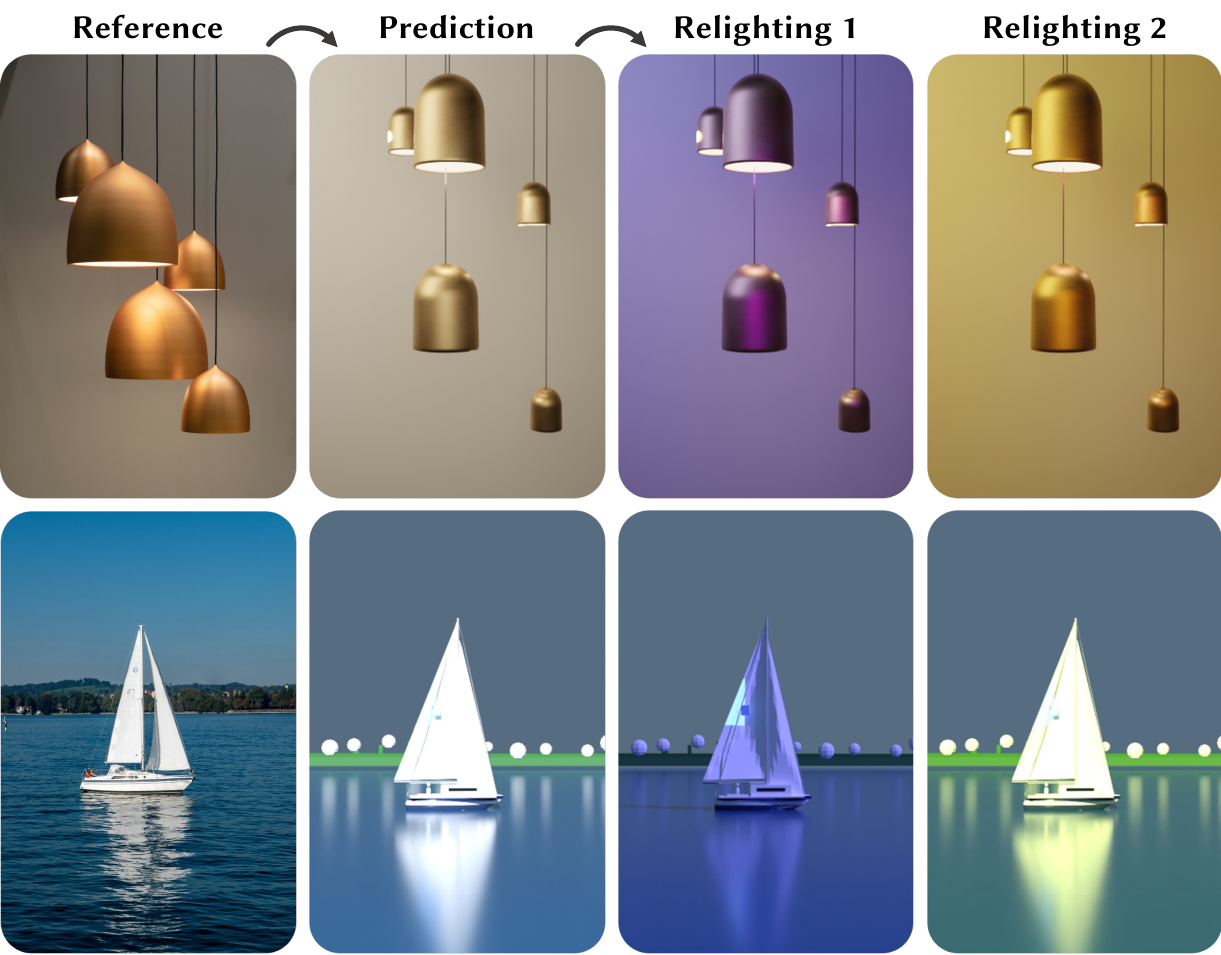}
    \vspace{-15pt}
    \caption{\textbf{Relighting.} Two reconstructed scenes (top: pendant lamps; bottom: sailboat) re-rendered under two new lighting configurations. Since lights are separately stored in Blender, new illumination can be applied by adding or reconfiguring light sources without re-running any part of the pipeline.}
    \Description{Two example scenes reconstructed by our pipeline and re-rendered under two synthetic lighting configurations each.}
    \vspace{-7pt}
    \label{fig:relighting}
  \end{figure}

\subsection{Applications}
\label{sec:applications}

A key advantage of producing an editable Blender file rather than an entangled latent representation is allowing immediate graphics operations without retraining or post-processing. In the following, we showcase examples on performing \textit{relighting}, \textit{object editing}, and \textit{physics simulation} on our reconstructed scenes.

\begin{figure*}[!p]
    \centering
    \begin{minipage}[c][\textheight][c]{\linewidth}
        \centering
        \includegraphics[width=\linewidth]{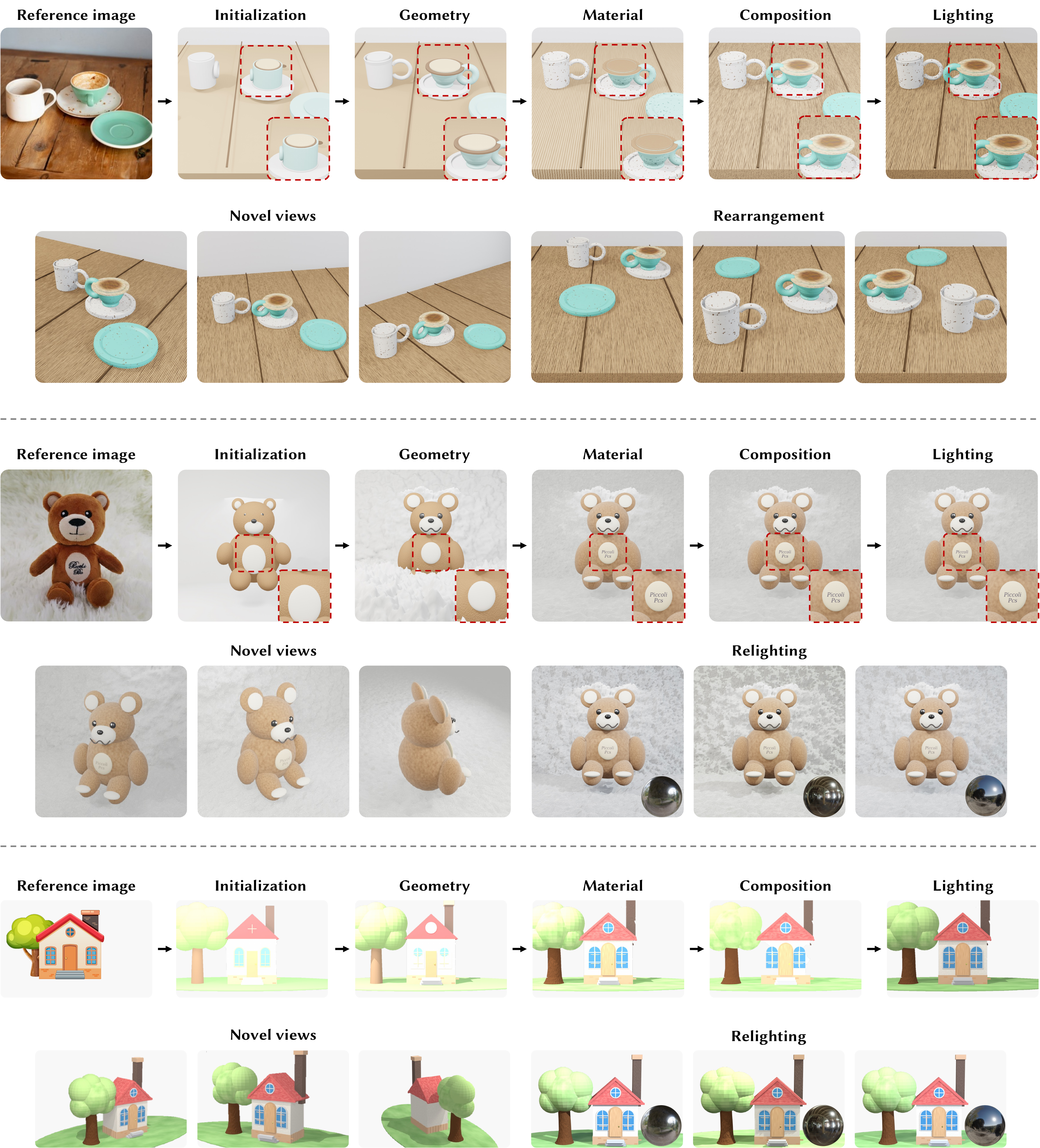}
            \caption{\textbf{Intermediate outputs across pipeline stages.} Two examples showing the rendered scene through our pipeline: starting from a coarse initial scaffold (\emph{Initialization}), through the four stages---\emph{Geometry}, \emph{Material}, \emph{Composition}, and \emph{Lighting}---each closed by its own generator--verifier loop, and the final image rendered from a VLM-determined camera (\emph{Camera-adjustment}, rightmost column). Each stage commits its output before the next begins, so every intermediate scene is itself a coherent, editable Blender program. This staged structure also underlies the downstream applications in Sec.~\ref{sec:applications}: materials, lights, and individual objects are exposed as separate, named stage outputs, so each can be modified directly without rerunning earlier stages---enabling the relighting, object editing, and physics simulation results that follow.}
        \Description{Gallery 2}
        \label{fig:intermediate}
    \end{minipage}
\end{figure*}

\smallskip\noindent\textbf{Relighting.}
Because geometry, materials, and light sources are committed as separate stage outputs (see the \emph{Lighting} column of Fig.~\ref{fig:intermediate}), lights can be added, removed, or reconfigured and the scene re-rendered under arbitrary new illumination without touching the rest of the recovered scene. The two examples in Fig.~\ref{fig:relighting} show our reconstructions re-rendered under two synthetic lighting configurations each, producing changes in dominant color, direction, and intensity while preserving the recovered geometry and materials.

\begin{figure}
    \centering
    \includegraphics[width=\columnwidth]{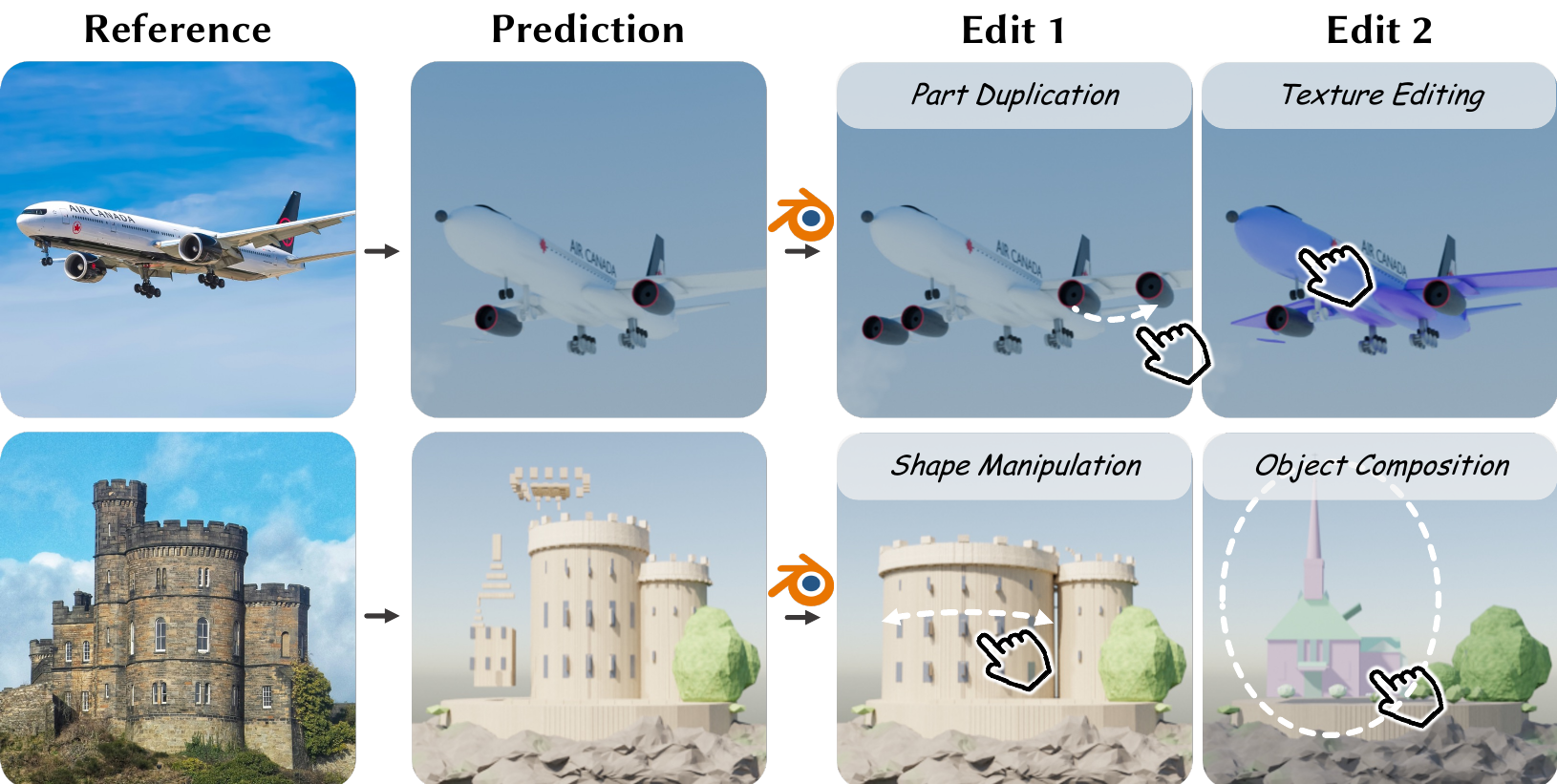}
    \vspace{-12pt}
    \caption{\textbf{Object editing.} Two reconstructed scenes (top: aircraft; bottom: castle), each shown alongside two example edits performed directly in Blender on the recovered scene graph: \emph{part duplication} and \emph{texture editing} for the aircraft; \emph{shape manipulation} and \emph{object composition} for the castle.}
    \Description{Four example object-level edits performed in Blender on two scenes reconstructed by our pipeline.}
    \label{fig:editing}
  \end{figure}

\smallskip\noindent\textbf{Object editing.}
The same structure makes per-object edits trivial: because each object is built independently in the Geometry and Material stages and only later assembled by the Composition stage, any node of the scene graph can be selected, moved, duplicated, retextured, or replaced directly in Blender. Fig.~\ref{fig:editing} shows four representative operations on two recovered scenes---\emph{part duplication} and \emph{texture editing} on an aircraft, and \emph{shape manipulation} and \emph{object composition} on a castle---each obtained by a small manual edit on the existing scene file. Fig.~\ref{fig:intermediate} additionally shows \emph{object rearrangement} on a recovered tabletop. None of these operations is straightforwardly available on the entangled latent representations produced by monolithic neural reconstruction methods.

\begin{figure}
    \centering
    \includegraphics[width=\columnwidth]{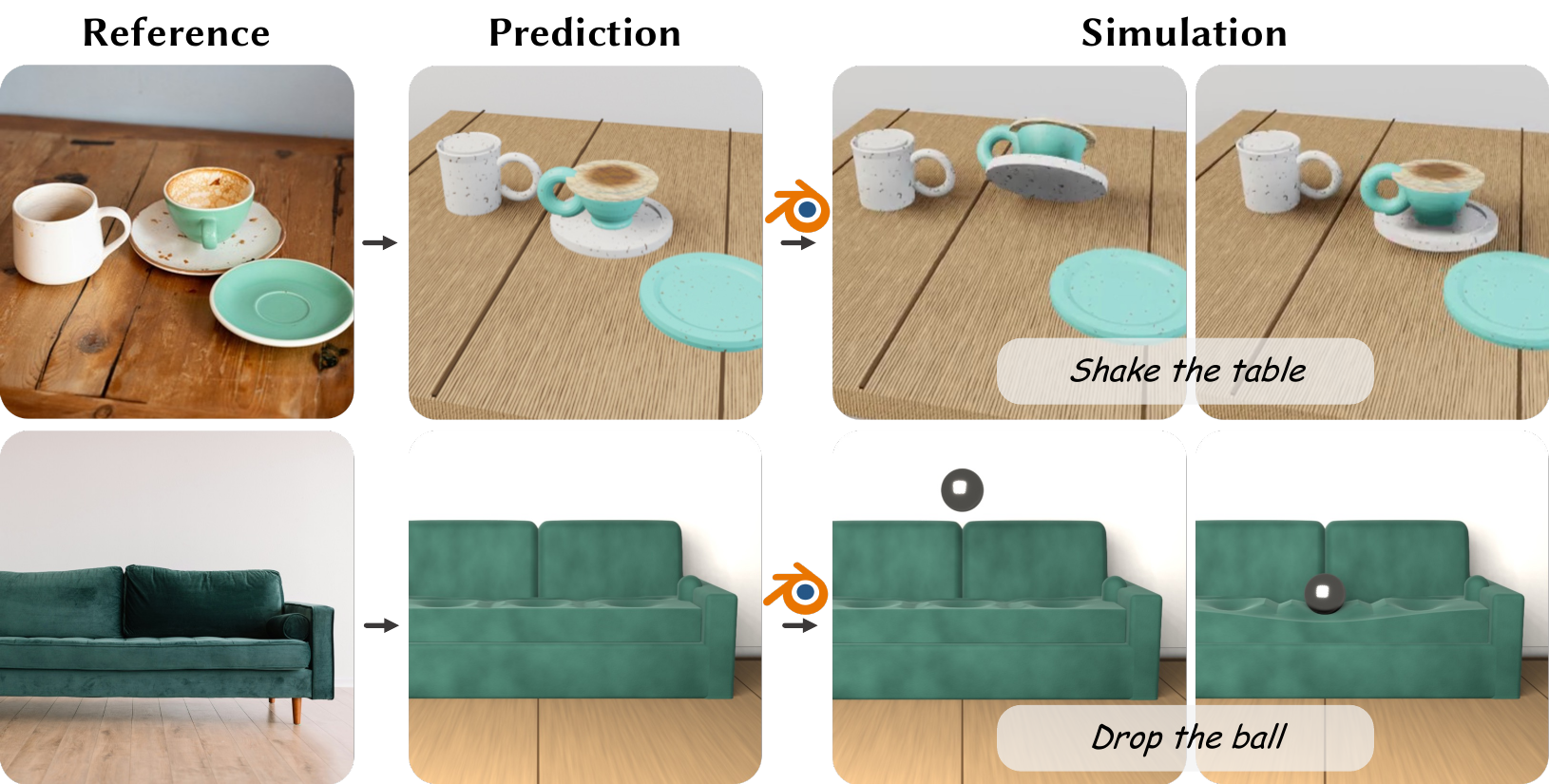}
    \vspace{-12pt}
    \caption{\textbf{Physics simulation.} Two reconstructed scenes used as input to Blender's built-in physics engine. Top: rigid-body dynamics---the recovered mugs and saucers slide and rattle when the table is given an external acceleration (\emph{shake the table}). Bottom: soft-body dynamics---a ball is dropped onto the recovered cushion, which deforms accordingly (\emph{drop the ball}). Both simulations run directly on the reconstructed scene graph in Blender.}
    \Description{Two reconstructed scenes used as input to Blender's physics engine, demonstrating rigid-body and soft-body simulation.}
    \label{fig:simulation}
  \end{figure}

\smallskip\noindent\textbf{Physics Simulation.}
Because the reconstructed scene is a structured collection of separately addressable meshes in Blender, it is also a valid input to Blender's built-in physics engine. Fig.~\ref{fig:simulation} shows two example scenarios run directly on our outputs: shaking the tabletop in the top row triggers rigid-body interactions among the recovered mugs and saucers, while dropping a ball onto the recovered couch in the bottom row exercises soft-body deformation of the cushion. Beyond attaching the appropriate Blender physics modifier (\emph{rigid body} or \emph{soft body}) to the relevant scene-graph nodes, neither scene requires remeshing, watertighting, or any other geometry repair---a direct consequence of the pipeline producing object-decomposed meshes rather than a single fused implicit representation, which would otherwise need to be converted into discrete physical entities before any simulation could be defined.

\subsection{Limitations}
\label{sec:limitations}
While our staged formulation enables more reliable executable inverse graphics reconstruction, errors introduced in early stages may propagate throughout the pipeline, leading to local minima from which later stages cannot easily recover. For example, inaccurate geometric reconstruction may constrain subsequent material, lighting, or compositional reasoning. One possible direction for mitigating this limitation would be to introduce additional global refinement passes that revisit and jointly optimize earlier scene factors after later reconstruction stages. However, such multi-round optimization would come at the expense of substantially increased computational cost and inference time.
Another notable limitation is the computational expense of repeated inference calls across multiple generator--verifier stages, resulting in substantially higher runtime and API cost compared to single-pass generation pipelines.

\section{Conclusion}
We introduced a staged executable inverse graphics framework that reconstructs editable Blender scenes directly from a single image using only a pretrained vision-language model, without task-specific training, specialized foundation models, or differentiable rendering. By decomposing reconstruction into a sequence of individually verifiable subproblems, each closed by its own generator--verifier stage, our framework enables the model to progressively recover geometry, materials, composition and lighting, while avoiding the entangled-output bottleneck that limits monolithic reconstruction pipelines. Future work could extend these ideas beyond static single-image reconstruction toward more challenging settings such as multi-image scene reconstruction, dynamic environments, physically grounded simulation, and long-horizon interactive editing. More broadly, our results suggest that staged executable scene representations provide a promising pathway for transforming increasingly capable general-purpose VLMs into controllable 3D content creation systems.

\clearpage

\clearpage
\bibliographystyle{ACM-Reference-Format}
\bibliography{main}

\end{document}